\documentclass{article}

\usepackage{times}
\usepackage{graphicx} 
\usepackage{subfigure} 

\usepackage{natbib}

\usepackage{amsmath}
\usepackage{bm}
\usepackage{amsthm}
\usepackage{float}
\usepackage{amsfonts}
\usepackage{graphicx}
\usepackage{wrapfig} 
\usepackage{caption}

\usepackage{algorithm}
\usepackage{algorithmic}

\usepackage{hyperref}

\usepackage[accepted]{icml2017}

\icmltitlerunning{Learning to Detect Sepsis}

\begin{document} 

\twocolumn[
\icmltitle{Learning to Detect Sepsis with a Multitask Gaussian Process RNN Classifier}



\icmlsetsymbol{equal}{*}

\begin{icmlauthorlist}
\icmlauthor{Joseph Futoma}{stats}
\icmlauthor{Sanjay Hariharan}{stats}
\icmlauthor{Katherine Heller}{stats}
\end{icmlauthorlist}

\icmlaffiliation{stats}{Dept. of Statistical Science, Duke University, Durham NC, USA}

\icmlcorrespondingauthor{Joseph Futoma}{jdf38@duke.edu}

\icmlkeywords{Multivariate Time Series, Gaussian Process, Recurrent Neural Network, Healthcare Application}

\vskip 0.3in
]

\printAffiliationsAndNotice{}  

\begin{abstract} 
We present a scalable end-to-end classifier that uses streaming physiological and medication data to accurately predict the onset of sepsis, a life-threatening complication from infections that has high mortality and morbidity.  Our proposed framework models the multivariate trajectories of continuous-valued physiological time series using multitask Gaussian processes, seamlessly accounting for the high uncertainty, frequent missingness, and irregular sampling rates typically associated with real clinical data.  The Gaussian process is directly connected to a black-box classifier that predicts whether a patient will become septic, chosen in our case to be a recurrent neural network to account for the extreme variability in the length of patient encounters.  We show how to scale the computations associated with the Gaussian process in a manner so that the entire system can be discriminatively trained end-to-end using backpropagation.   In a large cohort of heterogeneous inpatient encounters at our university health system we find that it outperforms several baselines at predicting sepsis, and yields 19.4\% and 55.5\% improved areas under the Receiver Operating Characteristic and Precision Recall curves as compared to the NEWS score currently used by our hospital.
\end{abstract} 

\section{Introduction}

Sepsis is a clinical condition involving a destructive host response to the invasion of a microorganism and/or its toxin, and is associated with high morbidity and mortality.  Without early intervention, this inflammatory response can progress to septic shock, organ failure and death \cite{sepsis}.  Identifying sepsis early improves patient outcomes, as mortality from septic shock increases by 7.6\% for every hour that treatment is delayed after the onset of hypotension \cite{sepsis-death}. It was also recently shown that timely administration of a 3-hour bundle of care for patients with sepsis (i.e. blood culture, broad-spectrum antibiotics, and a lactate measurement) was associated with lower in-hospital mortality \cite{nejm-sepsis}, further emphasizing the need for fast and aggressive treatment.  Unfortunately, early and accurate identification of sepsis remains elusive even for experienced clinicians, as the symptoms associated with sepsis may be caused by many other clinical conditions \cite{Jones}.
 
Despite the difficulties associated with identifying sepsis, data that could be used to inform such a prediction is already being routinely captured in the electronic health record (EHR).  To this end, data-driven early warning scores have great potential to identify early clinical deterioration using live data from the EHR.  As one example, the Royal College of Physicians developed, validated, and implemented the National Early Warning Score (NEWS) to identify patients who are acutely decompensating \cite{NEWS}.  Such early warning scores compare a small number of physiological variables (NEWS uses six) to normal ranges of values to generate a single composite score.  NEWS is already implemented in our university health system's EHR so that when the score reaches a defined trigger, a patient's care nurse is alerted to potential clinical deterioration.  However, a major problem with NEWS and other related early warning scores is that they are typically broad in scope and were not developed  to target a specific condition such as sepsis, since many unrelated disease states (e.g. trauma, pancreatitis, alcohol withdrawal) can result in high scores. Previous measurements revealed 63.4\% of the alerts triggered by the NEWS score at our hospital were cancelled by the care nurse, suggesting breakdowns in the training and education process, low specificity, and high alarm fatigue.  Despite the obvious limitation of using only a small fraction of available information, these scores are also overly simplistic in assigning independent scores to each variable, ignoring both the complex relationships between different physiological variables and their evolution in time.

The goal in this work is to develop a more flexible statistical model that leverages as much available data as possible from patient admissions in order to provide earlier and more accurate detection of sepsis.  However, this task is complicated by a number of problems that arise working with real EHR data, some of them particular to sepsis.  Unlike other clinical adverse events such as cardiac arrests or transfers to the Intensive Care Unit (ICU) with known event times, sepsis presents a challenge as the exact time at which it starts is generally unknown.  Instead, sepsis is typically observed indirectly through abnormal labs or vitals, the administration of antibiotics, or the drawing of blood cultures to test for suspected infection.  Another challenging aspect of our data source is the large degree of heterogeneity present across patient encounters, as we did not exclude certain classes of admissions.  More generally, clinical time series data presents its own set of problems, as they are measured at irregularly spaced intervals and there are many (often informatively) missing values.  Alignment of patient time series also presents an issue, as patients admitted to the hospital may have very different unknown clinical states, with some having sepsis already upon admission.  A crucial clinical consideration to be taken into account is the timeliness of alarms raised by the model, as a clinician needs ample time to act on the prediction and quickly intervene on patients flagged as high-risk of being septic.  Thus in building a system to predict sepsis we must consider timeliness of the prediction in addition to other metrics that quantify discrimination and accuracy.

Our proposed methodology for detecting sepsis in multivariate clinical time series overcomes many of these limitations.   Our approach hinges on constructing an end-to-end classifier that takes in raw physiology time series data, transforms it through a Multitask Gaussian Process (MGP) to a more uniform representation on an evenly spaced grid, and feeds the latent function values through a deep recurrent neural network (RNN) to predict the binary outcome of whether or not the patient will become (or is already) septic.  Setting up the problem in this way allows us to leverage the powerful representational abilities of RNNs, which typically requires standardized inputs at uniformly-spaced intervals, for our irregularly spaced multivariate clinical time series.  As more information is made available during an encounter, the model can dynamically update its prediction about how likely it is that the patient will become septic.  When the predicted probability of sepsis exceeds a predefined threshold (chosen to maximize predefined metrics such as sensitivity, positive predictive value, and timeliness), the model can be used to trigger an alarm.  

We train our model with real patient data extracted from the Duke University Health System EHR, using a large cohort of heterogeneous inpatient encounters spanning 18 months.  Our experiments show that using our method we can reliably predict the onset of sepsis roughly 4 hours in advance of the true occurrence, at a sensitivity of 0.85 and a precision of 0.64.  The benefits of our MGP classification framework are also apparent as there is a performance gain of 4.3\% in area under the ROC curve and 11.1\% in area under the Precision Recall curve, compared to the results of training an RNN to raw clinical data without a Gaussian Process to smoothly interpolate and impute missing values.  Our overall performance is also substantially better than the most common early warnings scores from the medical literature, and in particular we perform significantly better than the NEWS score currently in use at our hospital.  These large gains in performance will translate to better patient outcomes and a lower burden on the overall health system when our model is deployed on the wards in the near future as part of a randomized trial.

\section{Related Works}

There is a large body of works on the development and validation of early warning scores to predict clinical deterioration and other related outcomes.  For instance, the MEWS score \cite{MEWS} and NEWS score \cite{NEWS} are two of the more common scores used to assess overall clinical deterioration.  In addition, the SIRS score for systemic inflammatory response syndrome was part of the original clinical definition of sepsis \cite{SIRS}, although other scores designed for sepsis such as SOFA \cite{SOFA} and qSOFA \cite{qSOFA} have been more popular in recent years.  A more sophisticated regression-based approach called the Rothman Index \cite{Rothman} is also in widespread use for detecting overall deterioration.  Finally, \cite{trews} used a Cox regression to predict sepsis using clinical time series data, although they do not account for temporal structure since they simply create feature and event-time pairs from raw data.  

There has been much recent interest within machine learning in developing models to predict future disease progression using EHRs. For instance, \cite{Schulam} developed a longitudinal model for predicting progression of scleroderma, \cite{Futoma} presented a joint model for predicting progression of chronic kidney disease and cardiac events, and \cite{CTHMM} proposed a continuous-time hidden Markov model for progression of glaucoma.  However, these models operate on a longer time scale, on the order of months to years, which is different for our setting that demands real-time predictions at an hourly level of granularity.   The recent works  of \cite{ForecastICU} and \cite{mihaela-nips} are more relevant to our application, as they both developed models using clinical time series to predict a more general condition of clinical deterioration, as observed by admission to the Intensive Care Unit.

Although there has been some past methodological work on classification of multivariate time series, most of these approaches rely on clustering using some form of ad-hoc distance metric between series, e.g. \cite{ts-clust}, and comparing a new series to observed clusters.  More similar to our work are several recent papers on using recurrent neural networks to classify clinical time series.  In particular, \cite{RNNtimeseries} used Long-Short Term Memory (LSTM) RNNs to predict diagnosis codes given physiological time series from the ICU, and \cite{RNN-HF} used Gated Recurrent Unit RNNs to predict onset of heart failure using categorical time series of diagnosis and procedural codes.  Lastly, on a different note \cite{RNN_ts_miss} also used a variant of Gated Recurrent Unit networks to investigate patterns of informative missingness in physiological ICU time series.  

There are several related works that also utilized multitask Gaussian processes in modeling multivariate physiological time series.  For instance \cite{MGP-phys1} and \cite{MGP-phys2} used a similar model to ours, but instead focused more on forecasting of vitals to predict clinical instability, whereas our task is a binary classification to identify sepsis early.  Finally, our end-to-end technique to discriminatively learn both the MGP and classifier parameters builds off of \cite{marlin2016}.  However, our focus is more applied and the setting is more involved, as our time series are multivariate, of highly variable length, and may contain large amounts of missingness.

 \section{Proposed Model}

\begin{figure}[ht!]
\begin{center}
\centerline{\includegraphics[width=0.9\linewidth]{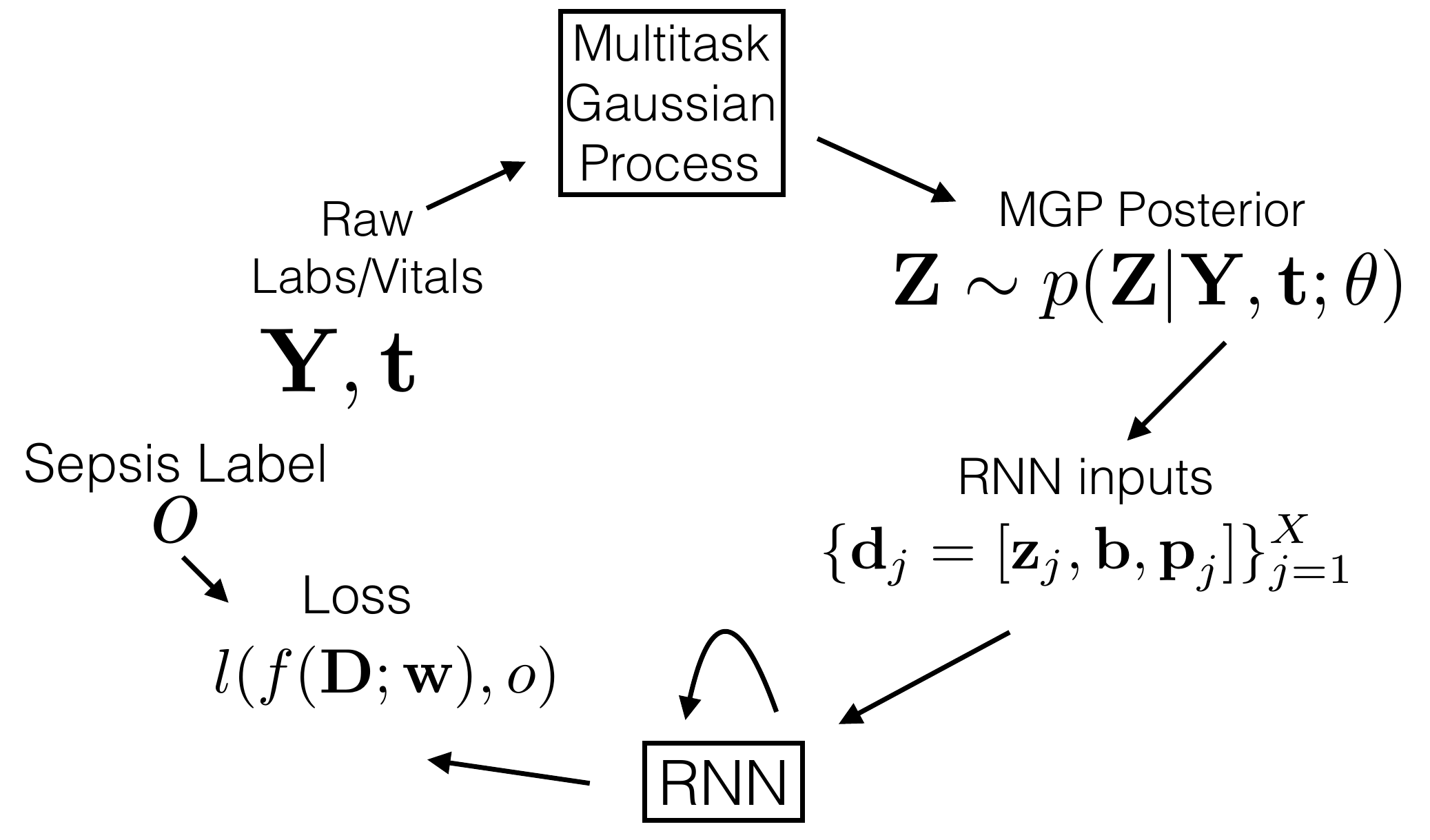}}
\caption{Model framework. Raw labs and vitals $\textbf{Y}$ at times $\textbf{t}$ are fed through a Multitask Gaussian Process to get imputed and interpolated values $\textbf{Z}$ at regularly spaced grid times $\textbf{x}$. These are fed into an RNN along with baseline covariates $\textbf{b}$ and medications $\mathcal{P}$ to predict the binary outcome $o$ (presence of sepsis).}
\end{center}
\end{figure}

We frame the problem of early detection of sepsis as a multivariate time series classification problem.  Given a new patient encounter, the goal is to continuously update the predicted probability that the encounter will result in sepsis, using all available information up until that time.  Figure 1 shows an overview of our approach.  We first introduce some notation, before presenting the details of the modeling framework, the learning algorithm, and the approximations to speed up learning and inference.  

\begin{figure}
\centering
\includegraphics[width=1\linewidth,height=7.5cm]{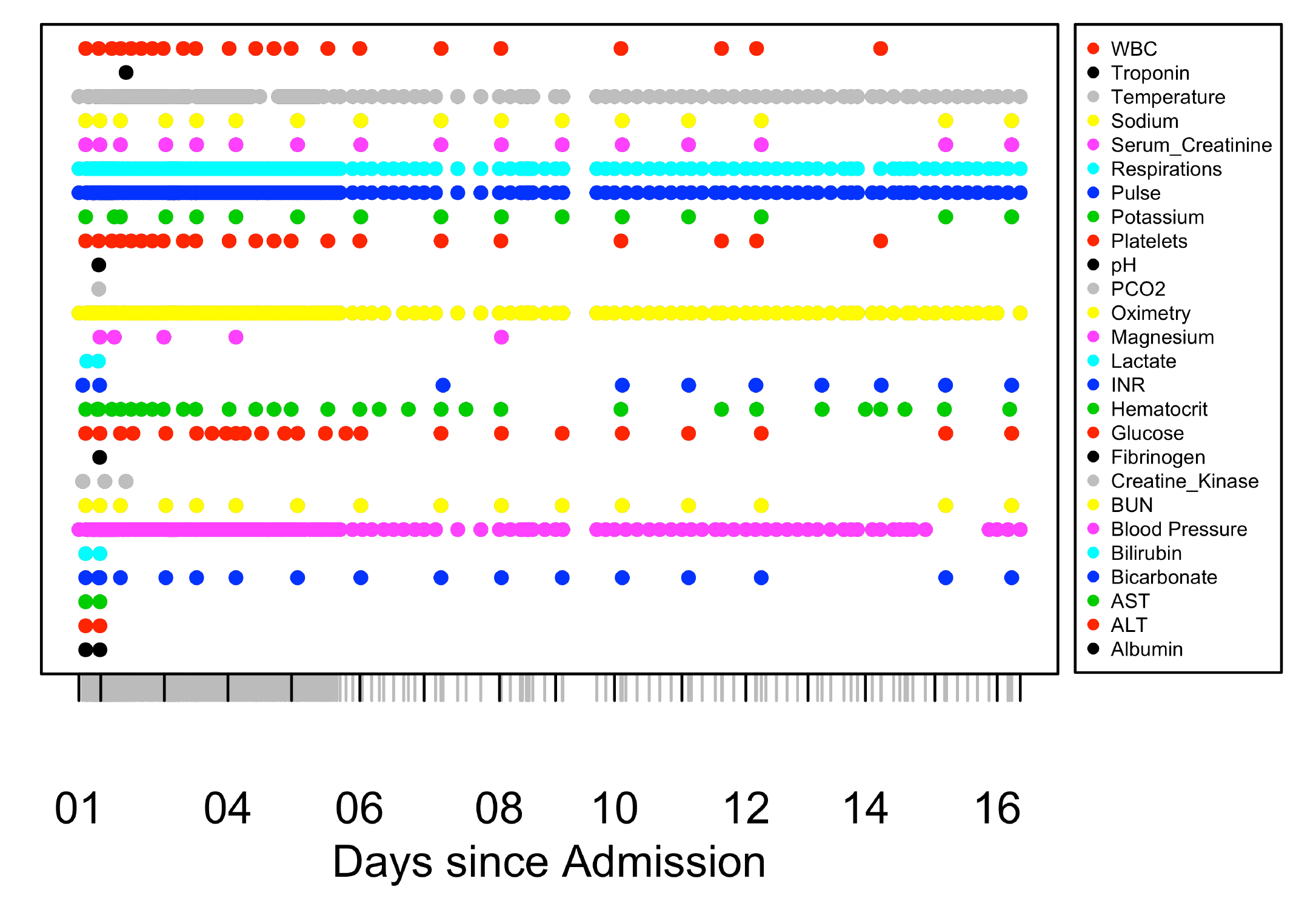}
\caption{An example patient encounter that shows when different lab and vital time series variables were measured, to highlight their irregular sampling rates. Not pictured are the 8 other physiological variables in our dataset that were never measured during this encounter (Ammonia, Bandemia, CK-MB, CRP, D-Dimer, ESR, LDH, PO2).}
\end{figure}

We suppose that our dataset $\mathcal{D}$ consists of $N$ independent patient encounters, $\{\mathcal{D}_i\}_{i=1}^N$.  For each patient encounter $i$, we have a vector of baseline covariates available upon admission to the hospital, denoted $\textbf{b}_i \in \mathbb{R}^B$, such as gender, age, and documented comorbidities.  At times $\textbf{t}_i = [t_{i1},t_{i2},\dots,t_{iT_i}]$ during the encounter we obtain information about $M$ different types of vitals and laboratory tests that characterize the patient's physiological state, where $t_{i1}=0$ is the time of admission.  These longitudinal values are denoted $\textbf{Y}_i = [\textbf{y}_{i1},\textbf{y}_{i2},\dots,\textbf{y}_{iM}] \in \mathbb{R}^{T_i \times M}$, with $\textbf{y}_{im} \in \mathbb{R}^{T_i}$ the vector of recorded values for variable $m$ at each time.  In practice, only a small fraction of this complete matrix is observed, since only a subset of the $M$ variables are recorded at each observation time.  We make no assumption about how long each encounter may last, so the length of the time series for each encounter is highly variable ($T_i \neq T_{i'}$) and these times are irregularly spaced, with each encounter having a unique set of observation times.  Additionally, during the encounter, medications of $P$ different classes are administered at $U_i$ different times (and it is possible for $U_i=0$).  We denote this information as $\mathcal{P}_i = \{(u_{i1},\textbf{p}_{i1}),(u_{i2},\textbf{p}_{i2}),\dots,(u_{iU_i},\textbf{p}_{iU_i}\}$, with $\textbf{p}_{ij} \in \{0,1\}^P$ a binary vector denoting which of the $P$ medications were administered at time $u_{ij}$.  This information is particularly valuable, because administration of medications provides some insight into a physician's subjective impression of a patient's health state by the type and quantity of medications ordered.  Finally, each encounter in the training set is associated with a binary label $o_i \in \{0,1\}$ denoting whether or not the patient acquired sepsis; we go into detail about how this is defined from the raw data in Section 4.1.  Thus, the data for a single encounter can be summarized as $\mathcal{D}_i = \{ \textbf{b}_i, \mathbf{t}_i, \mathbf{Y}_i, \mathcal{P}_i, o_i\}$.  

\subsection{Multitask Gaussian Processes}

Gaussian processes (GPs) are a common choice for modeling irregularly spaced time series as they are naturally able to handle the variable spacing and  differing number of observations per series.  Additionally, they maintain uncertainty about the variance of the series at each point, which is important in our setting since the irregularity and missingness of clinical time series can lead to high uncertainty for variables that are infrequently (or perhaps never) observed, as is often the case.  In order to account for the multivariate nature of our time series, we use a Multitask Gaussian Process (MGP) \cite{MGP}, an extension to GPs for handling multiple outputs at each time.  Let $f_{im}(t)$ be a latent function representing the true values of physiological variable $m$ for patient encounter $i$ at time $t$.  The MGP model places independent GP priors over the latent functions, with a shared correlation function $k^t$ over time.  We  assume each function has a prior mean of zero, so that the data has been centered.  Then, we have:
\begin{align}
\text{cov}(f_{im}(t),f_{im'}(t')) &= K^M_{mm'}k^t(t,t') \\
y_{im}(t) &\sim \mathcal{N}(f_{im}(t), \sigma_m^2)
\end{align}
	where $y_{im}(t)$ is the actual observed value.  Equivalently, the likelihood for a fully observed multivariate time series of $M$ measurements at $T$ unique times is:
\begin{align}
\text{vec}(\textbf{Y}_i) \equiv \textbf{y}_i \sim \mathcal{N}(\bm{0}, \Sigma_i) \\
\Sigma_i = K^M \otimes K^{T_i} + D \otimes I,
\end{align}
where $\textbf{y}_i$ is a stacked vector of all $M$ longitudinal variables at the $T_i$ observation times, and $\otimes$ denotes the Kronecker product.  $K^M$ is a full-rank $M \times M$ covariance matrix specifying the relationships among the variables, crucially allowing information from more frequently sampled variables to help improve learning about the variables infrequently (or perhaps never) measured. $K^{T_i}$ is a $T_i \times T_i$ correlation matrix (the variance can be fully explained by $K^M$) for the observation times $\textbf{t}_i$ as specified by the correlation function $k^t$ with parameters $\eta$ shared across all encounters. In this work we use the Ornstein-Uhlenbeck (OU) kernel function, $k^t(t,t') = e^{-|t-t'|/l}$, with a single length-scale parameter $\eta=l$.  The OU kernel is useful for modeling noisy physiological data, as draws from the corresponding stochastic process are only first-order continuous \cite{GPML}.  Finally, $D$ is a diagonal matrix of noise variances $\{\sigma_m^2\}_{m=1}^M$.  In practice, only a subset of the $M$ series are observed at each time, so the $MT_i \times MT_i$  covariance matrix $\Sigma_i$ only needs to be computed at the observed values.  This model is known in geostatistics as the intrinsic correlation model, since the covariance between different variables and between different points in time is explicitly separated, and is a special case of the linear model of coregionalization \cite{geostats}.

The MGP can be used as a mechanism to handle the irregular spacing and missing values in the raw data, and output a uniform representation to feed into a black box classifier.  To accomplish this, we define $\mathcal{X}$ to be a set of evenly spaced points in time (e.g. every hour) that will be shared across all encounters.  For each encounter, we denote a subset of these points by $\textbf{x}_i = (x_{i1}, x_{i2}, \dots, x_{iX_i})$, so that $x_{ij} = x_{i'j}$ if both series are at least $x_{ij}$ hours long.  The MGP provides a posterior distribution for the $M \times X_i$ matrix $\textbf{Z}_i$ of latent time series values at the grid times $\textbf{x}_i$ within this encounter, while also maintaining uncertainty over the values.  If we let $\textbf{z}_i = \text{vec}(\textbf{Z}_i)$, this posterior is also normally distributed with mean and covariance given by:
\begin{align}
\mu_{z_i} &= (K^M \otimes K^{X_i T_i }) \Sigma_i^{-1} \textbf{y}_i \\
\Sigma_{z_i} &= (K^M \otimes K^{X_i}) -  (K^M \otimes K^{X_i T_i}) \Sigma_i^{-1} (K^M \otimes K^{T_i X_i})
\end{align}
where $K^{X_i T_i}$ and $K^{X_i}$ are correlation matrices between the grid times $\textbf{x}_i$ and observation times $\textbf{t}_i$ and between $\textbf{x}_i$ with itself, as specified by the correlation function $k^t$.  The set of MGP parameters to be learned are thus $\bm{\theta} = (K^M, \{\sigma_m^2\}_{m=1}^M, \eta)$, and in this work we assume that they are shared across all encounters.  The structured input $\textbf{Z}_i$ can then serve as a standardized input to the RNN, where the raw time series data has been interpolated and missing values imputed.

\subsection{Classification Method}

We build off the ideas in \cite{marlin2016} to learn a classifier that directly takes the latent function values $\textbf{z}_i$ at shared reference time points $\textbf{x}_i = \{x_{ij}\}_{j=1}^{X_i}$ as inputs.  The time series for each encounter $i$ in our data can be represented as an MGP posterior distribution $\textbf{z}_i \sim N(\mu_{z_i}, \Sigma_{z_i}; \theta)$ at times $\textbf{x}_i$. This information will be fed into a downstream black box classifier to learn the label of the time series. 

Since the lengths of each times series are variable, the classifier used must be able to account for variable length inputs, as the size of $\textbf{z}_i$ and $\textbf{x}_i$ will differ across encounters.  To this end, we turn to deep recurrent neural networks, a natural choice for learning flexible functions that map variable-length input sequences to a single output.  In particular, we used a Long-Short Term Memory (LSTM) architecture \cite{LSTM}, as these classes of RNNs have been shown to be very flexible and have obtained excellent performance on a wide variety of problems.  At each time $x_{ij}$, a new set of inputs $\textbf{d}_{ij} = [\textbf{z}_{ij}^\top,\textbf{b}_i^\top,\textbf{p}_{ij}^\top]^\top$ will be fed into the network, consisting of the $M$ latent function values $\textbf{z}_{ij}$, the baseline covariates $\textbf{b}_i$, and $\textbf{p}_{ij}$, a vector of counts of the $S$ medications administered between $x_{ij}$ and $x_{i,j-1}$.  Thus, the RNN is able to learn complicated time-varying interactions among the static admission variables, the physiological labs and vitals, and administration of medications.  

If the function values $\textbf{z}_{ij}$ were actually observed at each time $x_{ij}$, they could be directly fed into the RNN classifier along with the rest of the observed portion of the vector $\textbf{d}_{ij}$, and learning would be straightforward.  Let $f(\textbf{D}_i;\bm{w})$ denote the RNN classifier function, parameterized by $\bm{w}$, that maps the $(M+B+P) \times X_i$ matrix of inputs $\textbf{D}_i$ to an output probability.  Learning the classifier given $\textbf{z}_i$ would involve learning the parameters $\bm{w}$ of the RNN by optimizing a loss function $l(f(\textbf{D}_i;\bm{w}), o_i)$ that compares the model's prediction to the true label $o_i$.  However, since $\textbf{z}_i$ is a random variable, this loss function to be optimized is itself a random variable.  Thus, the loss function that we will actually optimize is the expected loss $\mathbb{E}_{z_i \sim N(\mu_{z_i}, \Sigma_{z_i}; \theta)}[l(f(\textbf{D}_i;\bm{w}), o_i)]$, with respect to the MGP posterior distribution of $\textbf{z}_i$.  Then the overall learning problem is to minimize this loss function over the full dataset: 
\begin{align}
\bm{w}^*, \bm{\theta}^* = \text{argmin}_{w,\theta} \sum_{i=1}^N \mathbb{E}_{z_i \sim N(\mu_{z_i}, \Sigma_{z_i}; \theta)}[l(f(\textbf{D}_i;\bm{w}), o_i)].
\end{align}
Given fitted model parameters $\bm{w}^*, \bm{\theta}^*$, when we are given a new patient encounter $\mathcal{D}_{i'}$ for which we wish to predict whether or not it will become septic, we simply take $\mathbb{E}_{z_{i'} \sim N(\mu_{z_{i'}}, \Sigma_{z_{i'}}; \theta^*)}[f(\textbf{D}_{i'}; \bm{w}^*)]$ as a risk score that can be updated continuously as more information is available.  This approach is ``uncertainty-aware'', as the uncertainty in the MGP posterior for $z_i$ is propagated all the way through to the loss function.  Variations on this setup exist by moving the expectation. For instance, moving the expectation inside the classifier function $f$ swaps the MGP mean vector $\mu_{z_i}$ in place of $\textbf{z}_i$ in the RNN input $\textbf{D}_i$.  This approach will be more computationally efficient but discards the uncertainty information in the time series, which may be undesirable in our setting of noisy clinical time series with high rates of missingness.

\subsection{End to End Learning Framework}

The learning problem is to learn optimal parameters that minimize the loss in (7).  Since the expected loss $\mathbb{E}_{z \sim N(\mu_{z}, \Sigma_{z}; \theta)}[l(f(\textbf{D};\bm{w}), o)]$ is intractable for our problem setup, we approximate the loss with Monte Carlo samples:
\begin{align}
\mathbb{E}_{z \sim N(\mu_{z}, \Sigma_{z}; \theta)}[l(f(\textbf{D};\bm{w}), o)] &\approx \frac{1}{S} \sum_{s=1}^S l(f(\textbf{D}_s; \bm{w}), o), \\
\textbf{D}_s = [\textbf{Z}_s^\top, \textbf{B}^\top, \textbf{P}^\top]^\top, \; \; \text{vec}(\textbf{Z}_s) &\equiv \textbf{z}_s \sim N(\mu_z, \Sigma_z; \bm{\theta})
\end{align}
where $\textbf{B}$ and $\textbf{P}$ are appropriately sized matrices of the baseline covariates and medication counts over time.

We need to compute gradients of this expression with respect to the RNN parameters $\bm{w}$ and the MGP parameters $\bm{\theta}$.  This can be achieved with the reparameterization trick, using the fact that $\textbf{z} = \mu_z + R \xi$, where $\xi \sim N(0,I)$ and $R$ is a matrix such that $\Sigma_z = RR^\top$ \cite{reparam}.  This allows us to bring the gradients of (8) inside the expectation, where they can be computed efficiently.  Rather than choose $R$ to be lower triangular so that it can only be computed in $\mathcal{O}(M^3X^3)$ time with a Cholesky decomposition, we follow \cite{marlin2016} and let $R$ be the symmetric matrix square root, as this leads to a scalable approximation to be discussed in Section 3.4.  Finally, we train our model discriminatively and end-to-end by jointly learning $\bm{\theta}$ with $\bm{w}$, as opposed to a two-stage approach that would first learn and fix $\bm{\theta}$ before learning $\bm{w}$. 
 
 \subsection{Scaling Computation with the Lanczos Method}
 
 The computation to both learn the model parameters and make predictions for a new patient encounter is dominated primarily by computing the parameters of the MGP in (5) and (6) and then drawing samples for $\textbf{z}$ from it, as these are of dimension $MX$ (where $X$ is the number of reference time points). To make this computation more amenable to large-scale datasets such as our large cohort of inpatient admissions, we use the Lanczos method to obtain approximate draws from large multivariate Gaussians.  
 
\begin{algorithm}
    \caption{Lanczos Method to approximate $\Sigma^{1/2} \bm{\xi}$}
    \begin{algorithmic}
        \STATE \textbf{Input}: covariance matrix $\Sigma$, random vector $\bm{\xi}$, $k$
        \STATE $\beta_1 = 0$ and $\bm{d}_0 = \bm{0}$
        \STATE $\bm{d}_1 = \bm{\xi}/||\bm{\xi}||$
        \FOR{$j=1$ to $k$}
        	 \STATE\hspace{1pt} $\bm{d} = \Sigma \bm{d_j} - \beta_j \bm{d_{j-1}}$
        	 \STATE\hspace{1pt} $\alpha_j = \bm{d}_j^\top \bm{d}$        	 
        	 \STATE\hspace{1pt} $\bm{d} = \bm{d} - \alpha_j \bm{d}_j$   
        	 \STATE\hspace{1pt} $\beta_{j+1} = ||\bm{d}||$   
        	 \STATE\hspace{1pt} $\bm{d}_{j+1} = \bm{d}/\beta_{j+1}$	 
        \ENDFOR
		\STATE $ \bm{D} = [\bm{d}_1,\dots,\bm{d}_k] $
		\STATE $ \bm{H} = $ tridiagonal$(\bm{\beta}_{2:k},\bm{\alpha}_{1:k},\bm{\beta}_{2:k})$
        \STATE \textbf{Return}: $||\bm{\xi}||\bm{DH}^{1/2}\bm{e}_1$ \; \; // $\bm{e}_1=[1,0,\dots,0]^\top$
    \end{algorithmic}
\end{algorithm}

Recall that to draw from a multivariate Gaussian requires taking the product $\Sigma_{z}^{1/2} \xi$, where $\Sigma_{z}^{1/2}$ is the symmetric matrix square root and $\xi \sim N(0,I)$.  We can approximate this product using the Lanczos method, a Krylov subspace approximation that bypasses the need to explicitly compute $\Sigma_{z}^{1/2}$ and only requires matrix-vector products with $\Sigma_z$.  The main idea is to find an optimal approximation of $\Sigma_{z}^{1/2}$ in the Krylov subspace $\mathcal{K}_k(\Sigma_{z}, \xi) = \text{span}\{\xi,\Sigma_{z}\xi,\dots,\Sigma_{z}^{k-1}\xi\}$; this approximation is simply the orthogonal projection of $\Sigma_{z}\xi$ into the subspace.  See \cite{Krylov} for more details on the use of Krylov methods for sampling multivariate Gaussians.  In practice, $k$ is chosen to be a small constant, $k << MX$, so that the $\mathcal{O}(k^3)$ operation of computing the matrix square root of a $k \times k$ tridiagonal matrix can effectively be treated as $\mathcal{O}(1)$.  The most expensive step in the Lanczos method then becomes computation of matrix-vector products $\Sigma_z \bm{d}$. To compute these we use the conjugate gradient algorithm, another Krylov method, and it usually converges in only a few iterations.  We also use conjugate gradient when computing $\mu_z$ in (5) to approximate $\Sigma^{-1} \textbf{y}$.  Importantly, every operation in both the Lanczos method (Algorithm 1) and the conjugate gradient algorithm are differentiable, so that it is possible to backpropagate through the entire procedure during training with automatic differentiation.  
 
 \section{Experiments}
  
 \subsection{Data Description}
 
Our dataset consists of 49,312 inpatient admissions from our university health system spanning 18 months, extracted directly from our  EHR.  After extensive data cleaning, there are $M=34$ physiological variables, of which 5 are vitals, and 29 are laboratory values (see Figure 2), and they vary considerably in the number of encounters with at least one recorded measurement.  At least one value for each of the vital variables is measured in over 99\% of encounters, while some labs (e.g. Ammonia, ESR, D-Dimer) are very rarely taken, being measured in only 2-4\% of encounters, with most of the rest falling somewhere in the middle.  There were $b=35$ baseline covariates reliably measured upon admission (e.g. age, race, gender,  whether the admission was a transfer or urgent, comorbidities upon admission).   Finally, we have information on $P=8$ medication classes, where these classes were determined from a thorough review of the raw medication names in the EHR.  The patient encounters range from very short admissions of only a few hours to extended stays lasting multiple months, with the mean length of stay at 121.7 hours, with a standard deviation of 108.1 hours. As there was no specific inclusion or exclusion criteria in the creation of this patient cohort, the resulting population is very heterogeneous and can vary tremendously in clinical status.  This makes the dataset representative of the real clinical setting in which our method will be used, across the entire inpatient wards. Before modeling we log transform all $M$ physiological time series variables to reduce the effect of outliers, and then center and scale all continuous-valued inputs into the model.

For encounters that ultimately resulted in sepsis, we used a well-defined clinical definition to assess the first time at which sepsis is suspected to have been present.  This criteria consisted of at least two consistently abnormal vitals signs, along with a blood culture drawn for a suspected infection, and at least one abnormal laboratory value indicating early signs of organ failure.  This definition was carefully reviewed and found to be sufficient by clinicians.   Thus each encounter is associated with a binary label indicating whether or not that patient ever acquired sepsis; the prevalence of sepsis in our full dataset was 21.4\%.

\subsection{Experimental Setup}

We train our method to 80\% of the full dataset, setting aside 10\% as a validation set to select hyperparameters and a final 10\% for testing.  For the encounters that result in sepsis, we throw away data from after sepsis was acquired, as our clinical goal is to be able to predict sepsis before it happens for a new patient.  For non-septic encounters we train on the full length of the encounter until discharge.  We choose the shared reference times $\mathcal{X}$ to be evenly spaced at every hour starting at admission, as clinically the desire is for a risk score that will refresh only once an hour.

We compared our method (denoted ``MGP-RNN'') against several baselines, including several common clinical scoring systems, as well as more complex methods.  In particular, we compared our model with the NEWS score currently in use at our hospital, along with the MEWS score and the SIRS score.  Each of these scores are based off of a small subset of the total variables available to our methods.  In particular, MEWS uses five, NEWS uses seven, and SIRS uses four. These clinical benchmarks all assign independent scores to each variable under consideration, with higher scores given for more abnormal values, although they each use different thresholds and different values. 

 \begin{figure*}
  \includegraphics[width=\linewidth,height=7.5cm]{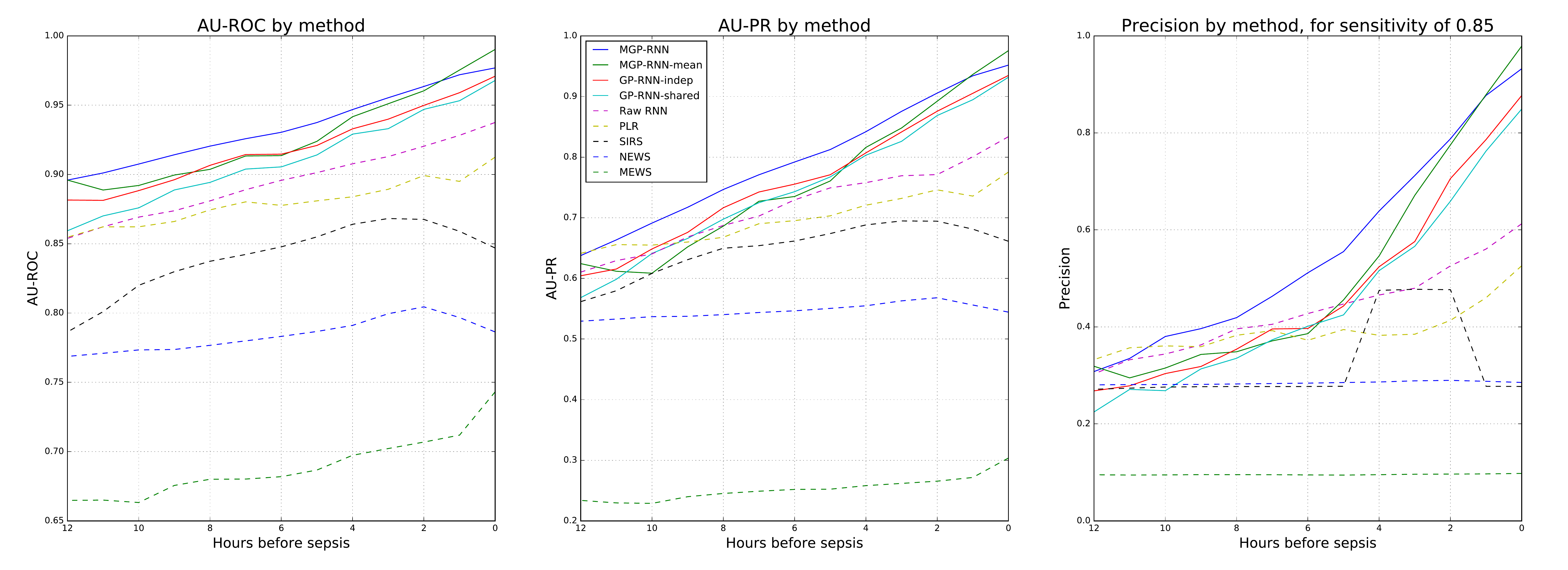}
  \caption{Left: Area under the Receiver Operating Characteristic curve for each method, as a function of the number of hours in advance of sepsis/discharge the prediction is issued (0-12 hours). Middle: Area under the Precision Recall curve as a function of time. Right: Precision as a function of time, for a fixed sensitivity of 0.85. Methods are all color coded according to middle legend; all GP/MGP-RNN methods are solid lines.}
  \vskip -0.2in
\end{figure*}

As a strong comparison to our end-to-end MGP-RNN classifier, we also trained an LSTM recurrent neural network from the raw data alone (denoted ``Raw RNN'' in Figure 3), with the same number of layers and hidden units as the network in our classifier (2 layers with 64 hidden units per layer).  The mean value for each vital and lab was taken in hourly windows, and windows with missing values carried the most recent value forward.  If there was no previously observed variable yet in that encounter, we imputed the mean.  In addition, we trained an $L_2$ penalized logistic regression baseline (``PLR'') using this imputation mechanism.

We also compare against a simplified version of the end-to-end MGP-RNN framework, (denoted ``MGP-RNN-mean'') where we replace the latent MGP function values $\textbf{z}$ with their expectation $\mu_{z}$ during both training and testing, to test the effect of discarding the extra uncertainty information.  Finally, to demonstrate the added value of using an MGP instead of independent GPs for each physiological variable, we trained two end-to-end GP-RNN baselines using (7), the same loss function as the MGP-RNN.  The first, ``GP-RNN-shared'', is equivalent to an MGP with $K^M = I$, i.e. no covariances across variables, and all variables share the same length-scale in the OU kernel.  The second, ``GP-RNN-indep'', is considerably stronger as it also has no covariances across variables, but allows each GP prior on the latent functions $f_m$ to have its own length scale in its OU kernel, i.e. $l_m \neq l_{m'}$.

To guard against overfitting we apply early stopping on the validation set, and use mild $L_2$ regularization for all RNN-based methods.  We train all models using stochastic gradient descent with the ADAM optimizer \cite{ADAM} using minibatches of 100 encounters at a time and a learning rate of $0.001$. To approximate the expectation in (8) we draw ten Monte Carlo samples.  We implemented our methods in Tensorflow\footnote{\url{https://github.com/jfutoma/MGP-RNN}}.  On a server with 63GB RAM and 12 Intel Xeon E5-2680 2.50GHz CPUs, the MGP-RNN method takes roughly 10 hours per epoch on the training set, and takes on average 0.3 seconds to evaluate a test case and generate a risk score.  All methods converged in a small number of epochs.

\subsection{Evaluation Metrics}

We use several different metrics to evaluate performance.  The area under the Receiver Operating Characteristic (ROC) curve (AU-ROC) is an overall measure of discrimination, and can be interpreted as the probability that the classifier correctly ranks a random sepsis encounter as higher risk than a random non-sepsis encounter.  We also report the area under the Precision Recall (PR) curve (AU-PR).  Importantly, we examine how these metrics vary as we change the window in which we make the prediction to see how far in advance we can reliably predict onset of sepsis.  
 
 \subsection{Results}
 
Our results show that the MGP-RNN classification framework yields clear performance gains when compared to the various baselines considered.  It substantially outperforms the overly simplistic clinical scores, and demonstrates modest gains over the RNN trained to raw data, the MGP-RNN-mean method that discards uncertainty information, and the univariate GP baselines.  

Figure 3 summarizes the results.  The four MGP/GP-RNN methods are in solid lines, and the other baselines are dashed. It is clear that these four methods perform considerably better than the other methods, especially in the last four hours prior to sepsis/discharge. 

The left and middle panes of Figure 3 display the AU-ROC and AU-PR for each method as a function of the number of hours in advance the prediction is made.  Generally the MGP-RNN performs best, followed by the three other MGP/GP-RNN baselines.  This is likely because it retains uncertainty information about the noisy time series (unlike MGP-RNN-mean), and can learn correlations among the different physiological variables to improve the quality of the imputation (unlike the GP-RNN methods).  As expected, the GP-RNN-indep baseline consistently outperforms the simpler GP-RNN-shared.  

The right pane of Figure 3 shows the tradeoff between precision and timeliness for a fixed sensitivity of 0.85 across the methods.  It is most useful to evaluate with such a high sensitivity as this is the setting clinicians typically want to use a risk score, in order not to miss many cases.  The MGP-RNN performs comparably to the MGP-RNN-mean within a few hours of sepsis, and demonstrates the biggest performance gains from about 3 to 7 hours beforehand.  Throughout, it has much higher precision than NEWS, MEWS, and SIRS, especially so in the few hours immediately preceding sepsis.  This is a very important clinical point, since clinicians want a method with very high precision and a low false alarm rate to reduce the alarm fatigue experienced with current solutions. Furthermore, being able to detect sepsis even a few hours early might substantially increase treatment effectiveness and improve patient outcomes.

\section{Conclusions and Clinical Significance}

We have presented a novel approach for early detection of sepsis that classifies multivariate clinical time series in a manner that is both flexible and takes into account the uncertainty in the series.  On a large dataset of inpatient encounters from our university health system, we find that our proposed method substantially outperforms strong baselines and a number of widespread clinical benchmarks.  In particular, our methods tend to have much higher precision, and thus they have much lower rates of false alarm.  For instance, at a very high sensitivity of 0.85 and when making predictions 4 hours in advance, there will be only roughly 0.5 false alarms for every true alarm generated by our approach, whereas for the NEWS score currently being used at our institution, there will be about 2.5 false alarms for every true alarm.  Thus, adoption of our method would result in a drastic reduction in the total number of false alarms.

In addition to the initial promise of our approach, there are a number of interesting directions to extend the proposed method to better account for various aspects of our data source.  In particular, we could incorporate a clustering component with different sets of MGPs for different latent subpopulations of encounters, to address high heterogeneity across patients.  The medication data might be better utilized to also learn the effect of medications on the physiological time series.  For instance, certain medications might have a sharp effect on certain vitals signs to help stabilize them; such treatment response curves could be learned observationally and applied to help improve predictions.  More sophisticated covariance structure in the multitask Gaussian process would allow for a more flexible model, since our assumption of a correlation function shared across all physiological streams may be overly restrictive.  Finally, use of additional approximations from the GP literature may further decrease the computational overhead and improve training times.

This work has the potential to have a high impact in improving clinical practice in the identification of sepsis, both at our institution and elsewhere. The underlying biological mechanism is poorly understood, and the problem has historically been very difficult for clinicians.  Use of a model such as ours to predict onset of sepsis would significantly reduce the alarm fatigue associated with current clinical scores, and could both significantly improve patient outcomes and reduce burden on the health system.  Although in this work our emphasis was on early detection of sepsis, the methods could be modified with minimal effort to apply to detection of other clinical events of interest, such as cardiac arrests, code blue events, admission to the ICU, and cardiogenic shock.  We are currently working to implement our methods directly into an application that can pull live data from our health system's EHR and present our model's predictions to a rapid response team.  This will allow us to apply our methods in a real-time clinical setting and their utility can be proven empirically, as data is collected on how accurate the alarms it raises are and how it is used on the actual wards.    
 
\section*{Acknowledgements} 
Joseph Futoma is supported by an NDSEG fellowship. Katherine Heller is supported by an NSF CAREER award.  This project was partially funded by the Duke Institute for Health Innovation.  Thanks to the other members of our team for their invaluable work on this project: Cara O'Brien MD,  Meredith Clement MD, Armando Bedoya MD, Mark Sendak MD, Nathan Brajer, and Bryce Wolery.

\small
\bibliography{CR_refs}
\bibliographystyle{icml2017}

\end{document}